\documentclass[11pt,a4paper]{llncs}
\usepackage{amsmath}
\usepackage{amssymb}
\usepackage{graphicx}
\usepackage{marvosym}
\usepackage{url}
\usepackage{fancyhdr}
\usepackage{comment}
\usepackage{geometry}
\newcommand{\keywords}[1]{\par\addvspace\baselineskip
\noindent\keywordname\enspace\ignorespaces#1}

\fancypagestyle{firstpage}{\fancyhf{}
}

\begin{document}

%\mainmatter  % start of an individual contribution

\title{\LARGE{Three Stage Narrative Analysis; Plot-Sentiment Breakdown, Structure Learning and Concept Detection}}

%\titlerunning{Lecture Notes in Computer Science: Authors' Instructions}

\author{\large{Taimur Khan  \and Ramoza Ahsan \and Mohib Hameed}}
\institute{\large{Department of Data Science and Artificial Intelligence, FAST National University of Computer and Emerging Science (FAST-NUCES)\\ Islamabad, Pakistan}}

%\author{Alfred Hofmann%
%\thanks{Please note that the LNCS Editorial assumes that all authors have used
%the western naming convention, with given names preceding surnames. This determines
%the structure of the names in the running heads and the author index.}%
%\and Ursula Barth\and Ingrid Haas\and Frank Holzwarth\and\\
%Anna Kramer\and Leonie Kunz\and Christine Rei\ss\and\\
%Nicole Sator\and Erika Siebert-Cole\and Peter Stra\ss er}

%\institute{Springer-Verlag, Computer Science Editorial,\\
%Tiergartenstr. 17, 69121 Heidelberg, Germany\\
%\url{http://www.springer.com/lncs}}


%\toctitle{Lecture Notes in Computer Science}
%\tocauthor{Authors' Instructions}

\maketitle

\thispagestyle{firstpage}

\begin{abstract}
Story understanding and analysis have been a challenging domain of Natural Language Understanding. The need for automated narrative analysis demands deep computational semantic captures, along with the syntactic analysis of the text. Moreover, a large amount of narrative data requires automated semantic analysis and computational learning rather than manual approaches to the analytical tasks. In this paper, we propose a framework that analyzes the sentiment arcs of movie scripts and performs an extended analysis regarding the context of the characters involved in the movie. The framework enables us to extract high and low concepts being delivered through the narrative. Using the methodologies of dictionary-based sentiment analysis, our proposed framework proceeded with a custom lexicon based sentiment analysis using LabMTsimple storylab module. The custom lexicon is based upon the Valence, Arousal, and Dominance scores (NRC-VAD lexicon). Furthermore, the framework advances the analysis by clustering similar sentiment plots using Ward’s hierarchical clustering technique. Our experimental evaluation using movie dataset demonstrates that the retrieved analysis is helpful to consumers and readers during the selection of a Narrative/Story. 

\end{abstract}

%\begin{abstract}
%The abstract should summarize the contents of the paper and should
%contain at least 70 and at most 150 words. It should be written using the
%\emph{abstract} environment.
%\keywords{We would like to encourage you to list your keywords within
%the abstract section}
%\end{abstract}

\section{Introduction}
Narratives are one of the main modes of human communication. It shapes how experiences and knowledge are communicated across cultures. From storytelling to novels and films, stories serve not only as entertainment, but also as a means of learning. Manual analysis of narratives in stories presents several challenges that researchers must navigate to ensure the accuracy and dependability of their findings. Some of the challenges include subjectivity and bias. Depending on the reader's background, experiences, and prejudices, stories can be interpreted in a variety of ways. Researchers may unintentionally project their own perspectives onto the data, impacting their interpretation of characters, themes, and events. Manual analysis of narratives in stories is a time consuming task that requires significant time and effort. Researchers have to allocate a generous amount of time for data analysis. It requires a deep understanding of the author's cultural context and other interpretations of the narrative. With the rise and advancements in artificial intelligence and natural language processing (NLP), it has become possible to study them computationally and with less effort. 

One key focus of computational narrative research is the way sentiment rises and falls over time, which can also be called the emotional arc. Many works of fiction conform to a limited set of recurring “story shapes”, while more recent work has extended these insights to film and television. At the same time, sentiment analysis methods have advanced beyond polarity detection, incorporating context and aspect-level preferences to produce finer representations. These developments emphasize the growing potential of automated tools to capture the emotional flow of narratives.

Equally important are the structural roles and conceptual layers that shape stories. Categorizing narrative components, such as tension, punishment, or victory, provides insight into plot progression. Distinctions between “high concept” and “low concept” are also beginning to be explored computationally through models that together analyze events and semantics. High-concept stories are defined by clear, exceptional assumptions that can be easily summarized in a sentence. It often relies on universal ideas. On the other hand, low-concept stories focus on subtle, character-driven, or culturally specific themes. The richness here comes from subtle interactions. By integrating high and low concepts, models can perform better and account for the variety of storytelling styles, and can provide more understanding of narratives.

\subsection{Motivation}
The research of the Genesis Story Understanding group at Massachusetts Institute of Technology (MIT) related to Natural Language Understanding has been a great help to Artificial General Intelligence \cite{GenesisSite}. Their Story learning projects are highly captivating, especially the ones under Concept learning and Story pattern learning \cite{finlayson2012learning}. Moreover the development of computational models on human story competence to develop accounts of human intelligence is a bleeding edge technology. The Genesis System has an interactive learning system called STUDENT, that takes in a small series of positive and negative examples of concepts and builds an internal model for these concepts \cite{barnwell2018using}. The impact of this research extends beyond MIT. It is providing valuable tools for other Artificial Intelligence communities, including the cognitive science community. The project helps researchers develop stronger models of narrative and concept understanding. It does this by formalizing and normalizing how stories and concepts can be learned and represented computationally. These contributions also support many practical applications, like automated story generation.

\subsection{Background}
Narrative structure here refers to the classification of a text segment into predefined action and noun categories such as villainy, punishment and difficulty etc.

Concept refers to the core idea of a story. Low concepts are simple and can come off as generic. However, these stories often contain more character development and nuance. Low concepts don’t have built-in conflicts and antagonists. Nor do they appear on their surfaces to be particularly unique or compelling. Here are some examples of low concepts: two teenagers fall in love; a widow struggles with grief; a detective solves a crime. 
 
High concepts pack a lot of punch in just a few words. They often wrestle with what-if questions and tend to contain built-in appeal while conveying a fresh or original idea, or a new twist on an old idea. Here are some examples of high concepts: What if scientists built dinosaurs from preserved DNA? (Jurassic Park); A lonely orphan is invited to a secret school for young wizards. (Harry Potter); What happens when artificial intelligence surpasses human intelligence? (The Terminator, The Matrix, Battlestar Galactica).

Most narratives fall into six categories, as outlined by \cite{Vonnegut1981}. The six story arcs are:

1. Rags to Riches: the protagonist rises in fortune or status.

2. Riches to Rags: the protagonist experiences a downfall or loss.

3. Man in a Hole: the protagonist faces difficulties but ultimately recovers.

4. Icarus: the protagonist rises and then falls.

5. Cinderella: the protagonist suffers, improves, and ends happily.

6. Oedipus: the protagonist suffers a downfall due to a flaw or fate.

The pattern and a structure of a narrative can be defined by a plot with sentiment based analysis against narrative timestamps. These plots are then clustered into what is known as six major story arcs. Furthermore, the human moods are psychologically triggered through the flow of a narrative. Thus, writers can move forward with their work by analyzing the story arc according to their flow and genre and can enhance the arcs during multiple draft reviews \cite{JoeBunting}.

Semantics are related to the meanings and inferences that can be extracted from a text. Natural Language Understanding mimics how human intelligence works in order to solve these problems, not only solving them but also devising efficient algorithms. Apart from this, Story Understanding is part of higher mental functionality. Human beings learn the best by imparting knowledge from stories. Hence to reach the levels of computational reasoning, the most initial steps are that stories and narratives should be analyzed for their structures and semantics. 

\subsection{Problem Statement}
\begin{figure}[h!]
\includegraphics[width=\linewidth]{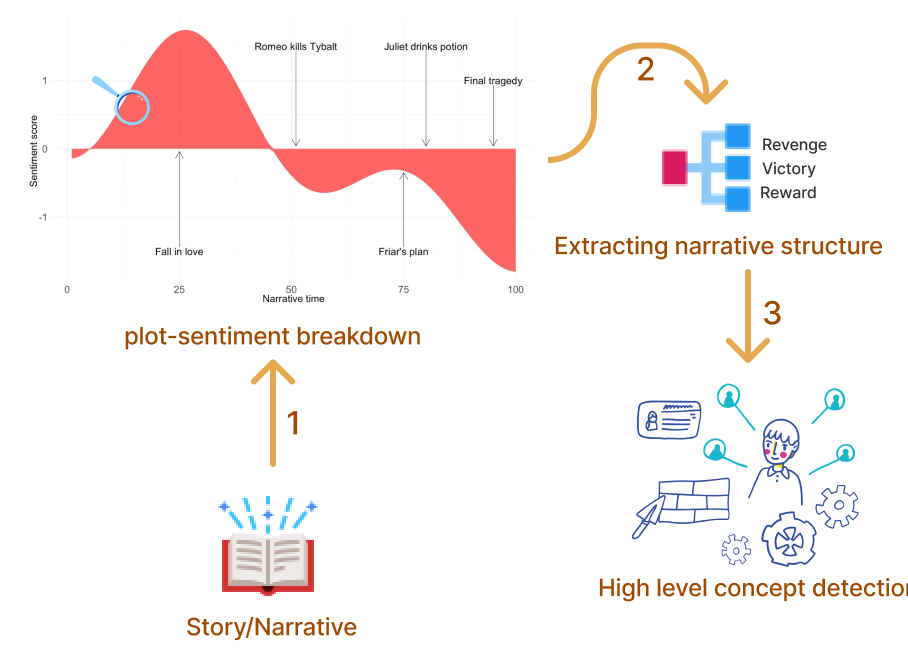}
\caption{Problem Breakdown: 3 Tier Narrative Analyzer; Plot-Sentiment Breakdown, Structure Learning and Concept Detection}
\label{problem}
\end{figure}

Given the significance of Story Analysis in Natural Language Understanding, we build on earlier studies on Story Arcs analysis and conduct a multistage analysis to determine how a movie script's pattern, flow, and structure are established. Moreover, we extract the sentiment plots and cluster them on the basis of emotional scores. The analysis allow us to compare our work and findings based on movie scripts with the existing works and solutions prepared regarding story arcs analysis on books and novels \cite{reagan2017towards}. Furthermore, we classify the narrative structure by analyzing the presence of known elements, such as reward, victory, or revenge, etc., and then extract high and low concepts from it. The main modules of our solution to the problem has been shown in Figure \ref{problem}.

The first key contribution of our work is the Plot-Sentiment Breakdown, where we applied a custom sentiment lexicon based on the NRC Valence, Arousal, and Dominance (VAD) model integrated into the LabMT framework \cite{labMT}. In contrast to conventional sentiment analysis methods that solely depend on positive or negative sentiments, our technique encompasses a broader spectrum.. It can generate multidimensional emotional attributes that allow for a better interpretation of the narrative flow. We also used segmentation and frequency-based scoring to improve this process. In this way, we are able to generate sentiment arcs for scripts that reflect dynamic emotional shifts within a story. Our results support previous research showing that stories often conform to a limited set of emotional trajectories. In addition to supporting previous research, we also expanded this analysis into the underexplored domain of movie scripts.

The second major contribution is the Structure Learning component, where we explored the classification of narrative segments into functional categories. Here, we classify the categories into tension, punishment, reward, and victory. This step moves into narrative semantics, providing a deeper structural understanding of story composition; moreover, the ability to label narrative units has effective implications for domains such as script writing and editing. Writers could use these tools to visualize the balance of narrative elements across their work.

The third stage is Concept Detection which is an integral component of our framework. As per our research, identification of high concepts versus low concepts plays an important role in the creative industries, as it often determines marketability and the impact it can make. While our current study primarily lays the groundwork by discussing the theoretical definitions of high and low concepts, we currently do not provide a computational model for automatically identifying these concepts, which remains a limitation of our work future work. We will implement computational methods for automatically identifying these categories based on semantic features, event structures, and novelty detection. Such research could potentially transform recommendation systems by suggesting content based on the conceptual depth.
\subsection{Related work}
\subsection{Segmentation}
In  \cite{barrow-etal-2020-joint} research, Barrow et al. focus on text segmentation by dividing the document into coherent structures. In previous works, document segmentation and segment labeling were dealt separately; however, in this approach, they aimed to address them jointly to produce better results. This paper introduces a new segment model named the Segment Pooling LSTM model, which efficiently performs the tasks together. 

As our project focuses on concept extraction from stories and narratives, it’s very important to specify the topics and semantics associated with each segment with utmost accuracy, hence, the idea to include this paper in our research was to have a distinct approach toward text segmentation. \cite{barrow-etal-2020-joint} observed that the segment boundaries and segment topics are highly dependent, and should be considered jointly, for accurately determining segment topics. This observation can further strengthen our research on high-level concept extraction from structures. The authors in this paper have proposed a neural model that jointly segments and labels the sentences, and maintains accuracy even on out-of-domain datasets, which would also help in our case and can be incorporated in our framework. The results show a 30\% decrease in segmentation error while improving segment labeling accuracy. This approach also works smoothly for both single and multi-label tasks, moreover, it generates better results as compared to all previous neural and non-neural models.

Segmentation is an important step in narrative analysis, as it determines how texts are broken down into units that can later be classified or assigned structural roles. Although traditional approaches typically rely on simple sequence-based models, recent research highlights the importance of capturing contextual dependencies between segments. For instance, \cite{zhang2020texting} introduced TextING, which uses a model that leverages both local word relationships and document-level graphs. Especially aimed at classification, their framework shows how these connections can improve the detection and labeling of narrative boundaries. Moreover, \cite{lee2024neuronarratives} proposed NeuroNarratives, a transformer-based system that learns narrative roles and story arcs simultaneously. By integrating segmentation with role assignment, their approach shows that accurately recognizing segment boundaries can enhance tasks such as structural role labeling and prediction of story arcs.

\subsection{Sentiment Analysis}
In research \cite{chen-etal-2020-aspect}, Chen et al. propose a methodology for Aspect Sentiment Classification (ASC). Existing research on Aspect Sentiment Classification is widely available, however, they always deal with sentence-level classification and document-level preference information independently. This paper discusses the importance of these two factors and proposes a different approach, in which they explore two kinds of document sentiment preference information (1) contextual sentiment consistency and (2) contextual sentiment tendency. 

The contextual sentiment consistency states that all the sentences in a document that have the same aspect, belong to the same sentiment polarity for that aspect. On the other hand, contextual sentiment tendency assumes that all the sentences in a document have the same sentiment polarity on all related aspects. These assumptions also make sense, and on this basis, the authors propose a new model called the Cooperative Graph Attention Networks (CoGAN). The approach uses two graphs to deal with both cases. The results show that this approach outperforms all state-of-the-art. Moreover, it also shows the importance of incorporating both intra-aspect consistency and inter-aspect tendency information for the Aspect Sentiment Classification task. Its effectiveness could help us extract accurate and realistic sentiments from our document, making our sentiment plots accurate and to-the-point, which would analyze important information from the text and build high-level concepts.

Recent research has started to look beyond analyzing single sentences or short text pieces. It is now focusing on how sentiment changes over the course of an entire story. For example, \cite{balestri2025multiagent} developed a multi-agent system that tracks narrative arcs in television series, and their approach shows how emotional patterns can be followed across multiple episodes, turning sentiment analysis into a continuous process. This idea connects closely with our work on film scripts, where emotions also unfold over time.

Graph-based methods have also advanced this field. \cite{zhang2020texting} introduced a model called TextING, which uses graph neural networks to capture the relationships between words and segments. The model was designed for text classification, but it is also useful for sentiment analysis because it considers how different parts of a text are connected. This kind of relational modeling can make sentiment predictions more accurate, specifically when separate sentences do not show the intended emotion.

\subsection{Story Arcs}
In Chapter 3 of \cite{reagan2017towards}, Reagen et al. propose that there are six basic shapes that dominate emotional arcs of stories. Firstly, the open access project Gutenberg corpus dataset has been used; roughly 50000 books were filtered to obtain a collection of 1327 English works of fiction.  Secondly a robust sentiment analysis tool is used to extract the reader-perceived emotional content of written stories; to generate a sentiment score, a dictionary based approach (LabMT dictionary) is taken for transparency and understanding of sentiment. 

After this, the emotional arcs are being analyzed using 3 methods independently; Matrix decomposition by Singular Value Decomposition (SVD) which finds the underlying basis of all of the emotional arcs, supervised learning by agglomerative (hierarchical) clustering with Ward’s method which classifies the emotional arcs into distinct groups, and unsupervised learning by a Self Organizing Map (SOM, a type of neural network) that generates arcs from noise which are similar to those in the corpus using a stochastic process.

Finally The first 6 SVD results(modes) agreed with the results of the machine learning and hierarchical clustering, and hence limited the results to these. Thus forming a broad support for the following six emotional arcs: Rags to riches (rise), Tragedy (fall),  Man in a hole (fall-rise), Icarus (rise-fall), Cinderella (rise-fall-rise) and Oedipus (fall-rise-fall).

\subsection{Structure Analysis and Text Classification}
In this paper \cite{zhang-etal-2020-every}, Zhang et al. propose an inductive text classification model named TextING using GNN. This paper discusses the limitations of native graph-based text classification models as they neither capture contextual word relationships within documents nor fulfill inductive learning of new words. The paper proposes a model in which the GNN can render detailed word to word relations using only training documents and generalize to new test documents. The model is such that it constructs a graph for a textual document using word co-occurrences and embeddings in the N-dimensional space; secondly, the model learns and updates the embeddings of word nodes by gathering information from its neighbors and then merging with its own representation. The experimental evaluations in the paper focus on model performance on unseen words during training and interpretability of the model on how words impact a doc. 

The results show that graph-based approaches outperform all other models and individual document graphs are better than global ones; secondly, the proposed model also performs better in inductive conditions when documents in training data are reduced; finally the model also shows a positive correlation between sentiment prediction and attention weights, which highlight important words, thus performing well in sentiment analysis.

\subsection{Semantics and Concept Extraction}
In this paper \cite{peng-etal-2017-joint}, Peng et al. focus on understanding the sequence of events that build up a story. Although it’s one of the challenging tasks in natural language understanding, the authors have proposed a different solution to this problem. The paper emphasizes the importance of events, as they carry multiple aspects of semantics like actions, entities, and emotions, which all contribute to the meaning of a story. The way these events are interdependent also contributes to the concept/semantics of the story, otherwise, the intended meaning may not be extracted accurately.

The authors have proposed: The frames, Entities, and Semantics Language Model (FES-LM), which jointly deals with the important aspects of the story’s semantics knowledge. They have also pointed out the three most important aspects: frames, entities, and semantics. The model is built from a plain training corpus with automatic annotation tools, which requires no human effort. The results prove the quality of the semantic language model. The model generates better results as compared to other word-level models. As our proposed work is related to the extraction of concepts from stories, we need to consider the sequence of events as they highly contribute to the meaning, understanding, and concept of a story. This paper could help us improve our understanding of the story in a depth-wise manner.

\section{Proposed Work}

\begin{comment}
\begin{itemize}
\item Bilinear: For all S, T $\in G_1$ ,  $e(aS,bT)=e(S,T)^{ab}$.

\item Non-degenerate: There exists $S$ and $T\in G_1$ such that
$e(S,T)\neq 1$.

\item Computable: There is an efficient algorithm to compute $e(S,T)$
for all $S,T\in G_1$. 
\end{itemize}
We have the following assumptions:
\begin{itemize}

\item The Decisional Diffie-Hellman problem(DDHP) in $G_1$ should be easy.
\item The DDHP in $G_2$, the computational Diffie-Hellman problem(CDHP) and the
discrete logarithm problem (DLP) in both $G_1$ and $G_2$ should be hard.
\item The inversion of the bilinear pairing be hard,
i.e., the bilinear pairing inversion problem(BPIP) is
defined as:\par

\end{itemize}
\end{comment}
We implement a three-tier Narrative Analyzer to capture emotional, structural, and conceptual aspects of stories. First, the Plot-Sentiment Breakdown outlines the story’s emotional arc. Next, the Structure Learning model classifies segments like reward, tension, or victory. Lastly, Concept Detection identifies high and low level story concepts. Together, these offer an organized approach to computational story understanding. 

\begin{figure}[h!]
\includegraphics[width=\linewidth]{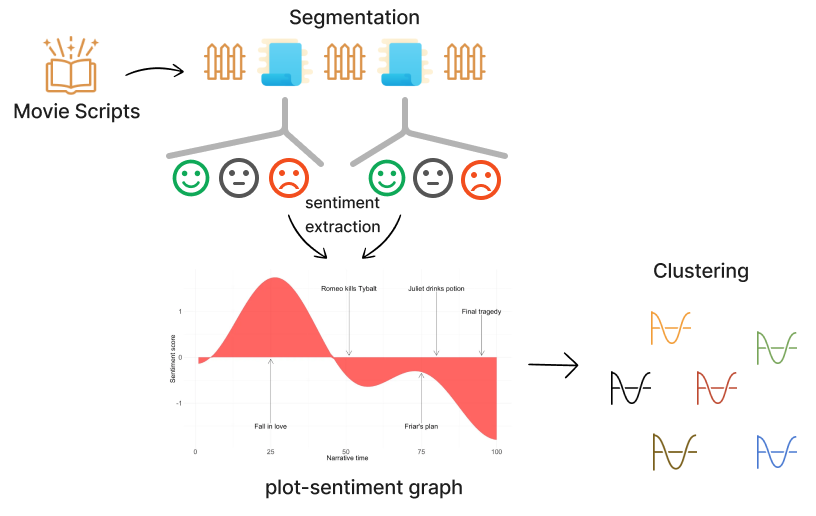}
\caption{Plot-Sentiment Breakdown for determining the Story Arc}
\label{breakdown}
\end{figure}

A \textbf{Plot-Sentiment Breakdown} for determining the Story Arc: As illustrated in Figure \ref{breakdown} the narrative is broken down into well structured segments by extracting the segment boundaries. We perform sentiment analysis in order to extract the sentiment/mood score of the narrative segment. A graph of the sentiment scores along the vertical axis is plotted against narrative timestamps and segments along the horizontal axis. An ML model subsequently classifies the resulting sentiment graph into one of the six major story arcs: Rags to Riches, Riches to Rags, Man in a Hole, Icarus, Cinderella, and Oedipus \cite{Vonnegut1981}.

\begin{figure}[h!]
\includegraphics[width=\linewidth]{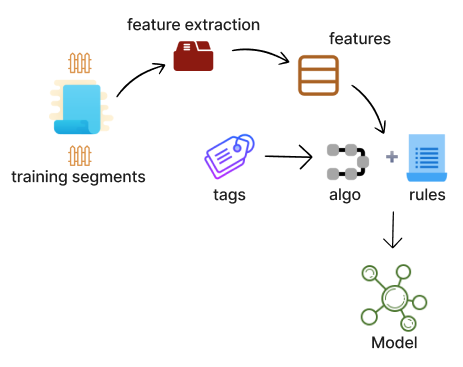}
\caption{Structure Learning and Text Classification: training procedure}
\label{classification_training}
\end{figure}

\begin{figure}[h!]
\includegraphics[width=\linewidth]{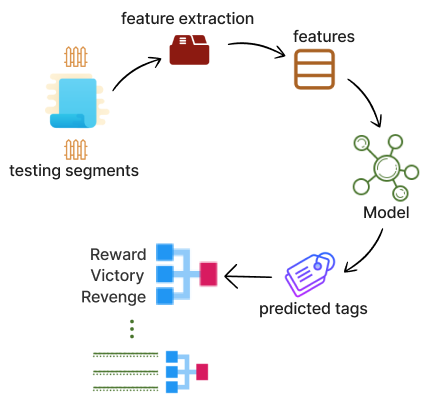}
\caption{Structure Learning and Text Classification: testing procedure}
\label{classification_testing}
\end{figure}

A \textbf{Structure Learning model} [Figure \ref{classification_training}, \ref{classification_testing}]: identifies what each segment is about, performing tasks such as topic labeling and structural categorization. Understanding what a given segment is talking about, for example topic labeling and structural organization of data. To extract the structure of a segment, a text classification model will be used. There will be predetermined classes or categories (e.g. reward, victory, tension, punishment) that a given text segment could fall into. 

\textbf{Concept Detection}: Semantic extraction and concept detection is the final part of the analyzer. In this part the system extracts high or low concepts from the narrative. The work on this part has yet to be researched, as this part is our future endeavor.

Our methodology is composed of three major phases namely Custom Lexicon dictionary for LabMT emotion rating, Segmentation of Script and Text Frequency vector and Sentiment Analysis using LabMT.

The first stage focuses on extracting emotional dynamics from narratives. To achieve this, we developed a custom sentiment lexicon that improves the accuracy of emotional arc detection across scripts.
\subsection{Custom Lexicon dictionary for LabMT emotion rating}
We use a customised Valence, Arousal and Dominance scored lexicon dictionary (NRC VAD lexicon). As shown below in Table 1, we set the custom lexicon on the same pattern as the built-in lexicon of LabMT. In order to yield better results for this project we have used the arousal scores vector replacing the happiness scores vector which come by default with labMT package \cite{reagan2017towards}. 

After constructing the lexicon, in the next step, we divided the movie scripts into meaningful narrative segments. This proper segmentation ensures that sentiment variations are analyzed in context.

\begin{table}[h!]
  \begin{center}
    \caption{Customized NRC-VAD lexicon}
    \label{tab:table1}
    \begin{tabular}{l|c|c|c|c} % <-- Alignments: 1st column left, 2nd middle and 3rd right, with vertical lines in between
      \textbf{Word} & \textbf{Ranking} & \textbf{Arousal} & \textbf{Valence} & \textbf{Dominance}\\
      \hline
    aaaaaaah & 1 & 0.606 & 0.479 & 0.291\\
    aaaah & 2 & 0.636 & 0.520 & 0.282\\
	aardvark & 	3 & 0.490 & 0.427 & 0.437\\
	aback & 4 & 0.407 & 0.385 & 0.288\\
	abacus & 5 & 0.276 & 0.510 & 0.485\\
	... & ... & ...	& ... & ...\\
	zoo & 20003 & 0.520 & 0.760 & 0.580\\
	zoological & 20004 & 0.458 & 0.667 & 0.492\\
	zoology & 20005 & 0.347 & 0.568 & 0.509\\
	zoom & 20006 & 0.520 & 0.490 & 0.462\\
	zucchini & 20007 & 0.321 & 0.510 & 0.250\\
     
    \end{tabular}
  \end{center}
\end{table}

\subsection{Segmentation of Script and Text Frequency vector}
The segmentation of script and generation of text fequency vector is composed of the following steps.
\begin{itemize}
\item Preprocess the movie script to extract words and remove all other characters and symbols. All words are then stored in a word list.
\item Define a window size which denotes number of words in a segment. 
\item For each segment do the following. \begin{itemize}
\item Concatenate the words equal to window size and form a segment string. 
\item Pass the segment string, arousal scores vector and VAD lexicon to the LabMT emotion function which will return a text frequency vector.
\end{itemize} 
\item Stack together text frequency vectors of all segments forming a matrix.
\end{itemize}

Once the text is segmented, we do sentiment analysis on each segment to generate emotion scores. This stage connects the lexicon and segmentation components and generates emotional representation.

\subsection{Sentiment Analysis using LabMT}
The following steps are followed in order to do sentiment analysis using LabMT.
\begin{itemize}
\item Define a context window that will help in accumulating text frequencies of the last 10 segments; this will propagate the influence of previous segments in the current segment sentiment score. This will also result in a much smoother plot.
\item Define an accumulated text frequency vector.
\item For each segment. \begin{itemize}
\item Sum the current text frequency vector with previous nine text freq vectors and update the accumulated text freq vector.
\item Remove stop words by zeroing their frequencies from the accumulated text freq vector.
\item Calculate the emotion score by passing stop words free accumulated text freq vector and arousal scores vector to the LabMT emotionV function which returns the emotion score.
\item Subtract the previous tenth text freq vector from the accumulated text freq vector so that at each point the accumulation is only holding a sum of most recent 10 segment text frequencies.
\end{itemize}
\item Store emotion scores of all segments
\end{itemize}

\section{Experimental Evaluation}
\subsection{Initial Clustering Mechanism}
For our experiments, we use a dataset of 1,000 movie scripts from publicly available repositories of film screenplays \cite{reagan2017towards}. Each script contains the dialogue, scene descriptions, and character actions. This allows for both sentiment and structural analysis. The dataset contains multiple genres. We use the Ward's method of hierarchical clustering which proceeds by minimizing variance between clusters of movie scripts\cite{reagan2017towards}. 
For the project we use the Scipy library's hierarchical clustering and fcluster modules.

\begin{figure}[h!]
\includegraphics[width=\linewidth]{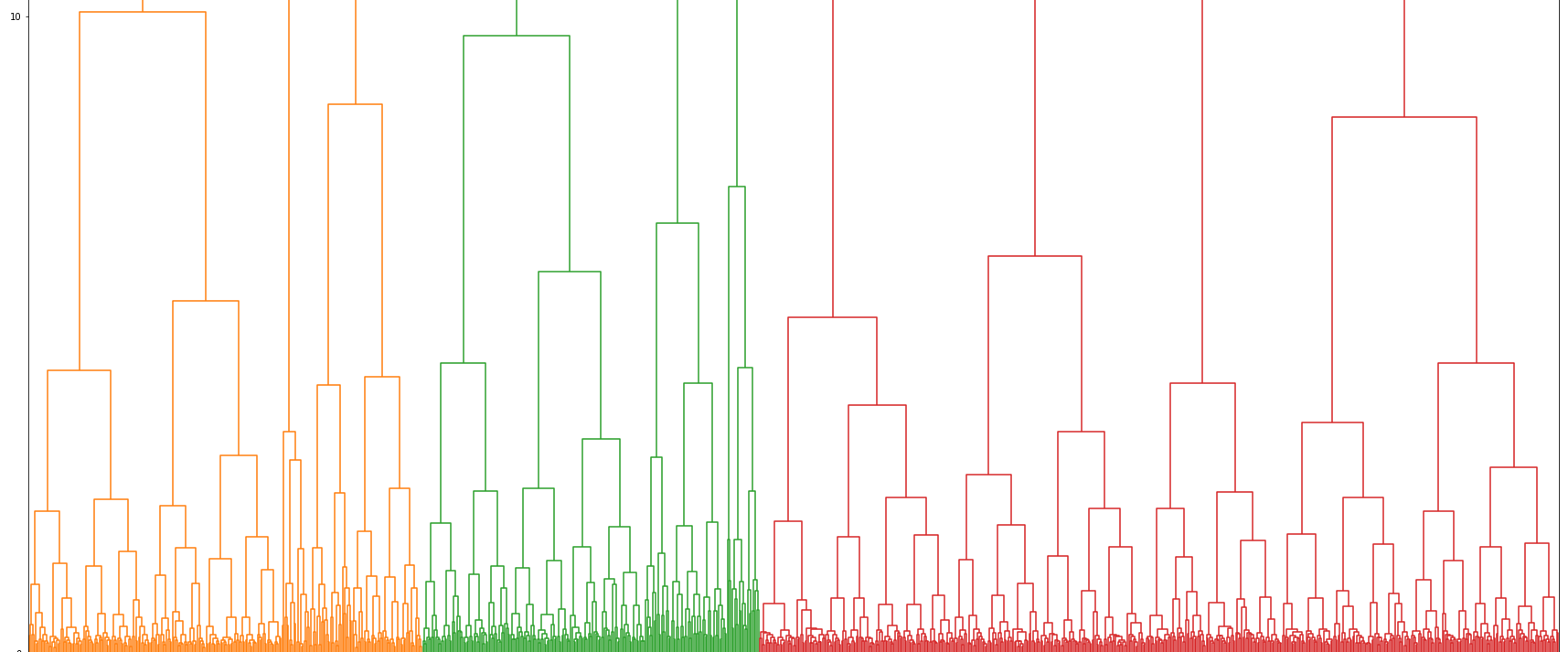}
\caption{Dendrogram showing hierarchical clustering of 1,000 movie scripts using Ward’s linkage. The vertical axis represents the distance (variance) between sentiment trajectories, and the horizontal axis groups scripts with similar emotional arcs. Three primary clusters were identified, indicating recurring emotional patterns across different film genres.}
\label{Dendogram}
\end{figure}

The hierarchical clustering dendogram was able to successfully generate three broad clusters as shown in Figure \ref{Dendogram}. In order to view the noisy graph plots in a clear representation, the setup was set to yield 100 smaller clusters for a thousand movie scripts dataset. 

\subsection{Results}
Our first experiments produced sentiment plots for a large number of movie scripts. Using Ward’s hierarchical clustering method, we grouped scripts that showed similar emotional patterns. The method produced three main groups, which suggested that very different movies can share common emotional shapes.

Looking at individual examples, we found that The Avengers, Blade Runner, and The Revenant all followed emotional curves that rose and fell in similar ways. This is interesting because the movies belong to different genres. When we divided the data into smaller clusters, more detailed differences appeared. It also showed that while many films follow a general shape, there are also unique details within certain genres.

To better see these similarities, we combined sentiment plots from scripts that belonged to the same cluster, and this reduced the noise seen in individual movies. Some clusters showed steady upward arcs, others sharp declines, and some were a mix. These results connect well with earlier theories that many stories follow a limited set of “universal” shapes. We did notice a few limitations as well. The raw sentiment plots were often noisy, and differences in script length made comparisons with others harder.

\begin{figure}[h!]
\includegraphics[width=\linewidth]{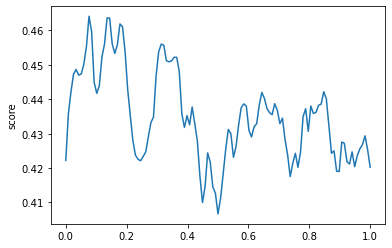}
\caption{Sentiment plot for The Revenant. The gradual rise–fall–rise pattern indicates phases of despair, struggle, and recovery. The similarity to previous plots underscores the clustering consistency among diverse narratives.}
\label{sentiment_Graph_Revenant}
\end{figure}

The experiment can be further analyzed by the differences between various sentiment graph plots of movie scripts as shown below in Figures \ref{sentiment_Graph_Avengers}, \ref{sentiment_Graph_Runner} and \ref{sentiment_Graph_Revenant}.

\begin{figure}[h!]
\includegraphics[width=\linewidth]{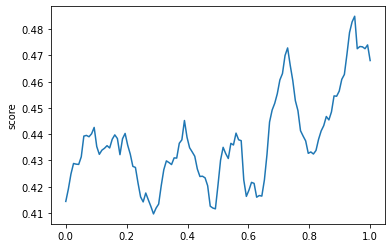}
\caption{Sentiment plot for The Avengers. The horizontal axis denotes narrative progression (segments), while the vertical axis represents sentiment or emotional score. The curve exhibits alternating peaks and troughs corresponding to the film’s major conflicts and resolutions.}
\label{sentiment_Graph_Avengers}
\end{figure}

\begin{figure}[h!]
\includegraphics[width=\linewidth]{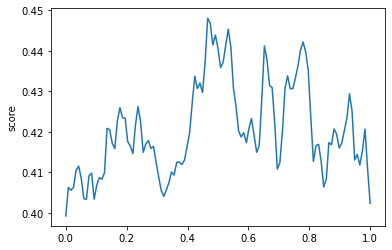}
\caption{Sentiment plot for Blade Runner. The plot follows the same axis convention as Figure 6, revealing a comparable emotional pattern that alternates between tension and reflection—supporting the hypothesis of universal story shapes.}
\label{sentiment_Graph_Runner}
\end{figure}

All of the above sentiment plots are somewhat similar to each other and according to our experiment and cluster results, these plots belong to a single cluster.

\begin{figure}[h!]
\includegraphics[width=\linewidth]{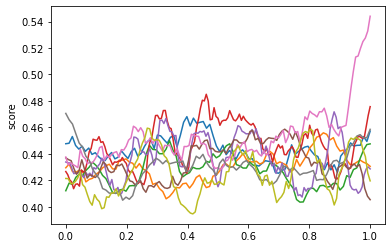}
\caption{Combined sentiment plot for Cluster No. 40, representing several scripts with upward emotional trajectories. Averaging within-cluster sentiment curves smooths local fluctuations and emphasizes the dominant “rags-to-riches” story shape.}
\label{cluster40}
\end{figure}

\begin{figure}[h!]
\includegraphics[width=\linewidth]{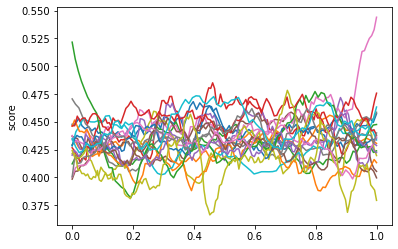}
\caption{Combined sentiment plot for Cluster No. 67. The curve shows alternating rises and declines that correspond to “Icarus” or “tragedy”-type arcs, reflecting narratives where optimism gradually transitions into loss.}
\label{cluster67}
\end{figure}

Figures \ref{cluster40} and \ref{cluster67} show the combined sentiment plot of multiple movie scripts belonging to particular smaller clusters.

\subsection{Analysis}
The sentiment emotion score vectors obtained in our experiments represent unprocessed outputs that have not yet undergone smoothing or additional transformation. The resulting plots might appear noisy, and this noise is somewhat a part of the segmentation process itself. As we know, when a script is divided into smaller units, small fluctuations in local word frequencies can cause sharp shifts in sentiment scores. While these shifts may reflect actual narrative dynamics at times, in many cases, it might not be true.

Another factor that might be influencing our results could be the script length and its variability. Movie scripts can vary dramatically depending on many factors. Some scripts are concise, while others are very detailed. This inconsistency makes direct comparison across scripts difficult. When sentiment arcs of unequal length are aligned on a common timeline, longer scripts appear more detailed, whereas shorter scripts look packed. This can distort outcomes and misrepresent emotional pacing. To overcome these limitations, we propose the following. The first involves the use of the Fourier Transform. By decomposing sentiment trajectories into frequency components, Fourier analysis enables us to separate underlying trends from high-frequency noise, and this would allow us to keep the global shape of a story’s emotional arc while filtering out local tips that are less meaningful for structural analysis. The second direction focuses on improving clustering strategies. Our current use of Ward’s hierarchical clustering provides a useful starting point by grouping scripts according to broad similarities in sentiment flow. However, this method alone does not preserve relationships among scripts. It also might not adapt well to the complex or even non-linear structure of data. To address this, we suggest combining hierarchical clustering with Self-Organizing Maps (SOMs). SOMs reduce dimensionality while preserving neighborhood relationships. This means that scripts with similar sentiment dynamics remain close together in the projection space. We could also generate clusters that are more accurate and also reflect global narrative emotion.

The benefits of these improvements are significant. At the same time, refining the analysis highlights the importance of future integration with structural and conceptual layers. Sentiment arcs alone, even when smoothed and clustered effectively, provide only some part of the information. To understand why some arcs resonate more strongly with audiences, we should also consider how specific structural roles and conceptual depth interact with these trajectories.
While our initial experiments demonstrate the feasibility of extracting sentiment arcs from movie scripts, they also reveal challenges. Some are related to noise and some to clustering accuracy. By integrating Fourier analysis and adopting hybrid clustering methods that combine hierarchical approaches with SOMs, I am sure we can significantly improve both the clarity of our findings.

The wider implications of this research are noteworthy. The work contributes to the scientific study of narratives, bridging the gap between computational methods and the psychology of storytelling. The applications of this research can also be noted. Publishers and content creators could use narrative analysis to provide new recommendation metrics. This can help audiences choose stories not just by genre or rating, but by their favorite emotions and conceptual type. Writers and creators could adopt narrative visualizations to strengthen emotional engagement in their drafts as well.

Our research is not without limitations: lexicon-based sentiment analysis is transparent, but it cannot capture the full context of human emotions. The structural classification relies on pre-defined categories that may overlook hybrid narrative functions. Moreover, concept detection remains only partially implemented. Our future work will therefore explore contextual embeddings to enhance sentiment scoring. We will further explore graph neural networks for more robust structural learning. Overall, this paper represents an initial step toward a comprehensive computational system for narrative analysis, a paper that integrates sentiment arcs, structural roles, and conceptual layers. By joining sentiment analysis, narrative theory, and machine learning, we move closer to a holistic model of story understanding. As a result of this research, we will be able to systematically analyze storytelling and generate insights of scientific and practical value. This will eventually serve as a useful tool for researchers, creators, and audiences, and help them understand creations of their own in a better way.
        
%\subsection{Security Analysis}
%
%The security of our scheme depends on the secure key distribution and the security of BLS signature. In \cite{blundo}, Blundo showed the use of symmetric t variable polynomial in degree k is k-secure. The theorem is as follows:
%        "In the scheme based on symmetric polynomial, if all coefficients
%of the symmetric polynomial in t variables of degree k are uniformly chosen in
%GF(q), then the t-conference key distribution scheme is k-secure, and optimal\cite{blundo}." This scheme is proven to be secure and also non-authorized entities had no way to obtain information of the secret key. On the other hand, the security of \cite{bls} Boneh et al. paper on short signatures from weil pairings is based on Computational Diffie-Hellman assumptions on certain elliptic curve. The theorem is as follows:"If \textbf{G} is a $(\tau, t', \epsilon' )$-GDH group, then the
%Gap Signature Scheme on \textbf{G} is $(t, q_H , q_S , \epsilon)$-secure against existential forgery on adaptive chosen-message attacks, where
%$t \leq t'-2\tau c_A (q_H + q_S )$
%and
%$\epsilon \geq 2e.q_S\epsilon' $,
%and $c_A$ is a small constant (in practice, at most 2)." The various parameters $\tau, t', \epsilon'..,etc$ are discussed in section $2.3$ and the proof of the theorem can be found in \cite{bls}. The security of our proposal follows immediately from these two results. A secure and authenticated channel for communication between the nodes have to be established in the first step in order to ensure the security of our scheme.                  

\section{Conclusion}
In this paper, we propose a three-stage narrative analysis framework that merges sentiment arc extraction, structural classification, and concept detection to analyze movie scripts. The motivation for this research originates from the fact that narratives play an important role in human cognition. Stories that shape communities have always been studied. Models that can capture the structures and sentiments of stories can be of great importance for our future in the domain of artificial intelligence.

Our experiments, in which we used a dataset containing over 1000 movie scripts, demonstrate the feasibility of this three-stage approach, and we also identified several coherent groups that aligned with known story arcs. Furthermore, case studies on scripts such as The Avengers, Blade Runner, and The Revenant illustrate how films with significantly different genres and tones can still show similar emotional trajectories. This also supports the idea of universal narrative structures or arrangements. Furthermore, our experiments also indicated limitations> One such example would be the presence of noise in sentiment plots due to variable script length and uneven segmentation. We have already identified possible solutions, including Fourier smoothing techniques to reduce noise and Self-Organizing Maps (SOMs) to keep topological structures in clustering. These advancements will form the basis of our future pipeline.

\section{Future Work}
While the present study establishes a foundation for narrative analysis, many directions are open for future research. The next most step we can take is to automate the Concept Detection stage. We plan to develop a computational model capable of identifying high and low concept narratives using semantic embeddings and novelty detection algorithms. This will transform the currently theoretical concept layer into a fully working module.

Another research can be to incorporate contextual sentiment models based on modern transformer architectures such as BERT, RoBERTa, and GPT-based encoders. These models can capture subtle emotional nuances, contextual dependencies, and character-driven expressions that traditional lexicon-based methods we have used often miss. Combining these can result in a more accurate semantic outputs.

Lastly, another future work potential can be to expand the data set to include multilingual and cross-cultural scripts. This will allow us to explore how emotional arcs and story structures may vary across different cultures. The proposed future work can advance the proposed work and help us understand stories in a better more effiecent way.
 
\nocite{*}
\bibliography{aim}
\bibliographystyle{acm}

\begin{comment}

\end{comment}
\vspace{2cm}

\section*{Authors}
\noindent {\bf Taimur Muhammad Khan} received MS in Data Science From Illinois Institute of Technology, IL, USA. He is currently working as Data Analyst and actively a part time instructor teaching AI to Healthcare professionals. His research interests include AI in Healthcare, NLP, and Data Analytics.\\

\noindent {\bf Ramoza Ahsan} received her PhD in Computer Science from Worcester Polytechnic Institute, MA, USA. She has served as an Assistant Professor in Artificial Intelligence and Data Science Department at National University of Computer and Emerging Sciences, Islamabad, Pakistan.  Her research interests
include data mining, big data analytics and machine
learning. She has also worked on association rule
mining and data integration problems.\\

\noindent {\bf Mohib Hameed} received BS in Artificial Intelligence from National University of Computer and Emerging Sciences, Islamabad, Pakistan. His research interests include AI in Healthcare and NLP.\\

\end{document}